\pdfoutput=1

\documentclass[11pt]{article}

\usepackage[]{acl2023}

\usepackage{times}
\usepackage{latexsym}

\usepackage[T1]{fontenc}

\usepackage[utf8]{inputenc}

\usepackage{microtype}

\usepackage{inconsolata}

\usepackage{graphicx}
\usepackage{amsfonts}
\usepackage{array}
\usepackage{multirow}
\usepackage{comment}
\newcolumntype{H}{>{\arraybackslash}m{2.25cm}}
\newcolumntype{M}{>{\arraybackslash}m{2.8cm}}
\newcolumntype{K}{>{\arraybackslash}m{2.5cm}}
\newcolumntype{G}{>{\arraybackslash}m{3.9cm}}
\newcolumntype{T}{>{\centering\arraybackslash}m{0.35cm}}
\newcolumntype{N}{>{\centering\arraybackslash}m{0.3cm}}
\newcolumntype{Z}{>{\centering\arraybackslash}m{1.5cm}}

\usepackage{caption}
\usepackage{subcaption}

\title{DialoGen: Generalized Long-Range Context Representation \\for Dialogue Systems}

\author{Suvodip Dey\textsuperscript{\textmd{1}}, Maunendra Sankar Desarkar\textsuperscript{\textmd{1}}, Asif Ekbal\textsuperscript{\textmd{2}}, P.K. Srijith\textsuperscript{\textmd{1}}\\
   \textsuperscript{1}Indian Institute of Technology Hyderabad, India \\ \textsuperscript{2}Indian Institute of Technology Patna, India \\
   \texttt{\small{suvodip15@gmail.com, maunendra@cse.iith.ac.in, asif@iitp.ac.in, srijith@iith.ac.in}} \\
   }

\begin{document}

\maketitle

\begin{abstract}
Long-range context modeling is crucial to both dialogue understanding and generation. The most popular method for dialogue context representation is to concatenate the last-$k$ utterances in chronological order. However, this method may not be ideal for conversations containing long-range dependencies, i.e., when there is a need to look beyond last-$k$ utterances to generate a meaningful response. In this work, we propose DialoGen, a novel encoder-decoder based framework for dialogue generation with a generalized context representation that can look beyond the last-$k$ utterances. The main idea of our approach is to identify and utilize the most relevant historical utterances instead of last-$k$, which also enables the compact representation of dialogue history with fewer tokens. We study the effectiveness of our proposed method on both dialogue generation (open-domain) and understanding (DST). Even with a compact context representation, DialoGen performs comparably to the state-of-the-art models on the open-domain DailyDialog dataset. We observe a similar behavior on the DST task of the MultiWOZ dataset when the proposed context representation is applied to existing DST models. We also discuss the generalizability and interpretability of DialoGen and show that the relevance score of previous utterances agrees well with human cognition.

\end{abstract}

\section{Introduction}
One of the key challenges in dialogue systems is modeling long-range context \cite{ds-book}. Human conversations can be lengthy and may contain long-range dependencies among turns. While having a conversation, we often refer back to names, topics, or other information that was mentioned long before the current dialogue turn. For example, Table \ref{tbl:example2} shows an open-domain conversation from the DailyDialog \cite{dailydialog} dataset. We can observe that in Turn 11, ``\textit{it}'' refers to the word ``\textit{hats}'', which is mentioned only once in the first turn. Understanding such long-range dependencies is critical for long-range context modeling, which can be beneficial for both dialogue generation and understanding.

\begin{table}[t]
\centering
\begin{small}
\begin{tabular}{|l|l|}
\hline
\textbf{Turn} & \textbf{Utterance} \\ \hline
1 & Oh , so many kinds of winter \underline{\textit{hats}} . \\ \hline
2 & What is your favorite color , miss ?\\ \hline
3 & Red . \\ \hline
4 & Here you are. It ' s very attractive . \\ \hline
5 & May I try it on ? \\ \hline
6 & Go ahead . \\ \hline
7 & Is there a mirror around here ? \\ \hline
8 & Right over there . \\ \hline
9 & Does it suit me ? \\ \hline
10 & Yes , you look very nice . \\ \hline
11 & How much is \underline{\textit{it}} ? \\ \hline
\end{tabular}
\caption{\label{tbl:example2} A sample conversation from DailyDialog}
\end{small}
\end{table}

The main challenge of dialogue context modeling comes from the fact that conversations can be arbitrarily long and complex in nature. To encode arbitrary long conversations, researchers started adapting a hierarchical recurrent encoder framework that contains an utterance-level and a dialogue-level encoder \cite{hred}. However, this approach cannot fully leverage the benefits of the utterance level features (discussed in Section \ref{sec:h_enc}). After the evolution of Transformers \cite{transformer}, the most popular approach to context modeling is to concatenate the historical utterances and use a transformer decoder (or encoder-decoder) model to generate the response. As the sequence length of a transformer is limited, people generally use only the last-$k$ utterances according to memory limit. Despite its simplicity, this method has produced state-of-the-art results for almost all kinds of dialogue-related tasks \cite{dialogpt, trippy, skt}. Since the existing dialogue datasets have a scarcity of long-range dependencies among turns, looking only at last-$k$ turns is enough to generate a good aggregate-level performance. Although this phenomenon of relying only on recent turns can be observed in short and simple real-world conversations, the same cannot be said for more complex scenarios.

In this work we propose \textbf{DialoGen}\footnote[1]{Code is available at \href{https://github.com/SuvodipDey/DialoGen}{github.com/SuvodipDey/DialoGen}}, an open domain \textbf{Dialo}gue system with \textbf{Gen}eralized context representation strategy. The primary objective of DialoGen is to enrich dialogue context modeling by addressing long-range dependencies such that arbitrarily long conversations can be handled in an easy and interpretable way. The central idea of our approach is to \emph{find the relevant historical utterances along with a vector representation of the entire context that can guide the generation of a meaningful response}. The main contributions of our work are as follows:
\begin{itemize}
    \item We propose DialoGen, a novel dialogue generation framework with a generalized representation for long-range dialogue context.
    \item The proposed context representation method can handle arbitrary long conversations and works even when the context for the current turn might have been presented much earlier in the conversation. The relevance scores over all the previous turns help to understand the long-range dependencies among dialogue turns, which enhances the generalization and interpretability of the context representation.
    \item DialoGen achieves comparable performance to state-of-the-art models on dialogue generation and understanding, even with its short and compact representation of dialogue history.
    \item Detailed discussion on the generalizability and interpretability of the proposed approach, along with a psycholinguistic perspective.
    
\end{itemize}

\section{Background and Related Works}
The existing neural network approaches for context modeling can be broadly categorized into two classes: Concatenation-based and  Hierarchical.

\subsection{Concatenation-based Encoding}
\label{sec:c_enc}
In this approach, historical utterances are concatenated to represent the context. In pre-Transformer era, the concatenation-based encoding strategy was a go-to method to train an RNN based encoder-decoder \cite{nmt} for dialogue generation \cite{rnn-conv}. A major issue with this approach is that the concatenated utterances can be very long, depending on the conversation. Moreover, modeling long-range dependencies with an RNN/LSTM is difficult. This is why researchers started switching to hierarchical encoders (Section \ref{sec:h_enc}) to handle long conversations. However, concatenation-based encoding again came to the forefront after the emergence of Transformer architecture \cite{transformer}. Most of the Transformer based dialogue models concatenate previous utterances and finetune the decoder on language modeling task \cite{tranfertransfo, dialogpt, plato, dialoflow, dialogved} to achieve state-of-the-art results on various dialogue datasets. Note that Transformers have a limit on maximum sequence length. This is why all these dialogue models can only take last-$k$ previous utterances as input based on a pre-defined maximum sequence length. Hence, they are not able to look beyond last-$k$ turns and thereby cannot capture very long-range dependencies among dialog turns. There are variations of Transformer (like Big-Bird \cite{bigbird}, Poolingformer \cite{poolingformer} etc.) that reduce computation complexity of self-attention operation from $O(n^2)$ to $O(n)$, enabling longer sequence length. However, looking at more context does not necessarily solve the problem of long-range dependencies, as there might still exist dependencies beyond the concatenated context that could be passed according to the maximum allowed sequence length.

 \begin{figure*}[t]
    \begin{center}
        \includegraphics[scale=0.82]{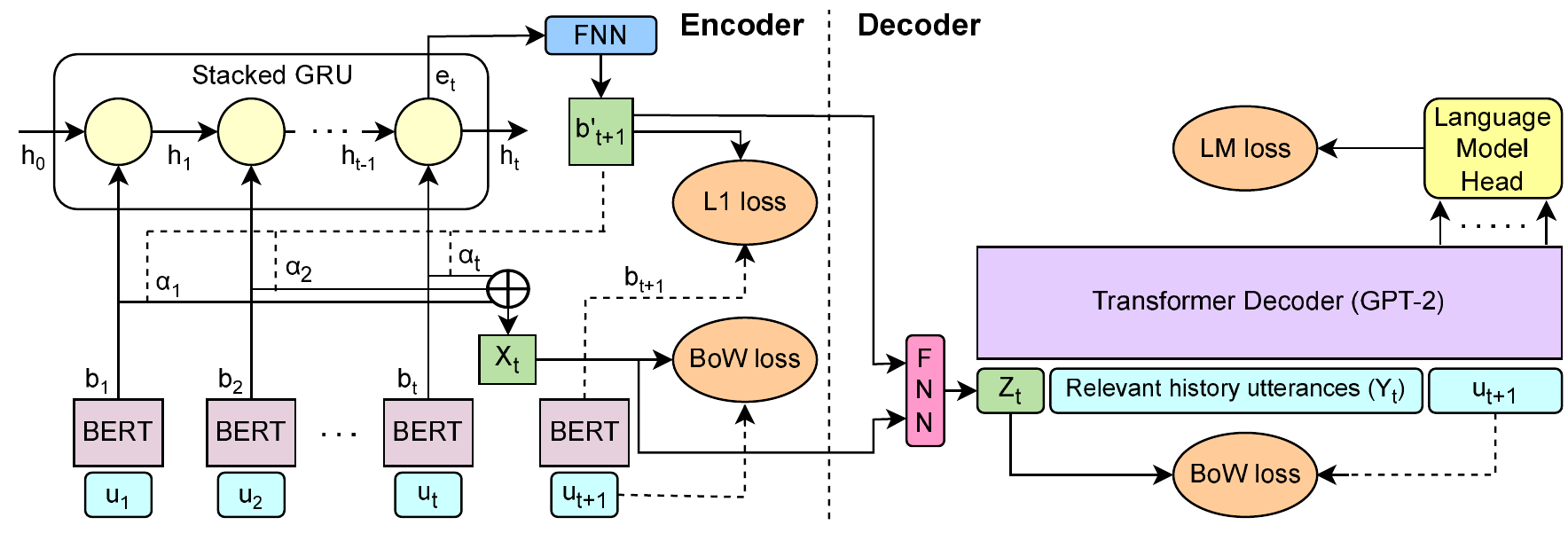}
    \caption{Architecture of DialoGen}
    \label{fig:arc}
    \end{center}
\end{figure*}

\subsection{Hierarchical Encoding}
\label{sec:h_enc}
In this strategy, the encoding of arbitrary long conversations is achieved through a hierarchical encoder. Each utterance is first encoded using an utterance-level encoder. The encoded utterances are then fed to a dialogue-level encoder to get the final context representation. As discussed in Section \ref{sec:c_enc}, vanilla RNN-based encoder-decoder architecture cannot handle long conversations. To address this issue, researchers started adopting hierarchical recurrent encoders where two separate RNN/LSTM are employed as the utterance-level and dialogue-level encoders. Models like HRED \cite{hred} and VHRED \cite{vhred} fall under this category. There are few works that use BERT as an utterance-level encoder \cite{skt}. \citet{incremental-transformer} proposed an Incremental Transformer for hierarchical recurrent encoding. DialogBERT \cite{dialog-bert} uses two separate BERT \cite{bert} encoders to realize hierarchical encoding. Although DialogBERT can handle lengthy conversations, theoretically, the number of turns is limited by the maximum sequence length of BERT. 
The main advantage of hierarchical encoding is its ease of encoding long conversations. However, these models depend only on the final context vector for response generation. Not considering word/token level features can fail to capture the complex correlation between all the words in the input and output sequences, which may be required for dialogue generation. Moreover, in real-world conversation, we often reuse words/phrases from past utterances in our replies for which word/token level features are important. The decoders of concatenation-based methods put attention on all the context tokens during response generation. This is why most of the state-of-the-art results are reported using concatenation-based encoding. Hierarchical encoding-based models like HRAN \cite{hran} and ReCoSa \cite{recosa} try to address this issue by additionally considering attention on utterance-level words/tokens. But doing so makes the dialogue generation dependent on context length, which again brings back some of the limitations discussed in Section \ref{sec:c_enc}.

\section{Methodology}
\label{sec:method}
In this section, we describe our proposed dialogue generation framework, DialoGen. Let $D = \{u_1, u_2, u_3, ...\}$ be a multi-turn conversation where $u_i$ represents the utterance at turn $i$. The objective of dialogue generation is to generate $u_{t+1}$ given $D_{\leq t}$ i.e. $\{u_1, u_2, u_3, ..., u_{t}\}$. The main idea of our approach is to combine the advantages of both concatenation-based and hierarchical encodings and provide a generalized context representation for dialogue systems that is adaptive to long-range dependencies. The framework is based on an encoder-decoder architecture, as shown in Fig. \ref{fig:arc}.

\subsection{Encoder}
DialoGen encoder is basically a hierarchical recurrent encoder with few added elements. At a given turn $t$, the encoder first predicts the encoding of the next response. This predicted encoding ($b_{t+1}'$) is then used to find a relevance score ($\alpha ^{(t)}$) for all the previous utterances. Finally, $\alpha^{(t)}$ is used to compute a vector representation ($X_t$) of the entire context such that the prediction of ground-truth words/tokens is maximized.

\textbf{Hierarchical Encoding}: We use BERT \cite{bert} and GRU (Gated Recurrent Unit) \cite{enc-dec} as our utterance-level and dialogue-level encoders respectively. At each turn $t$, the utterance-level encoder ($f_\phi$) takes $u_t$ as input and outputs $b_t$. Here $f_\phi$ is defined as the mean of all the tokens of the second-to-last layer of the BERT model. The utterance-level encoding is then passed to the stacked GRU ($g_{\psi}$) with $l$ layers to generate the contextual representation $e_t$. The procedure of obtaining the contextual representation can be summarized as,  
\begin{equation} \label{eqn:2}
    b_t = f_{\phi}(u_t) \in \mathbb{R}^{d}
\end{equation}
\begin{equation} \label{eqn:3}
    e_t, h_t = g_{\psi}(b_t, h_{t-1})
\end{equation}
where $d$ is the dimension of BERT embedding, $e_t \in \mathbb{R}^{d}$ is the output of the GRU and $h_t \in \mathbb{R}^{l \times d}$ is the GRU hidden state. The initial hidden state $h_0$ is set to a zero matrix.

\textbf{Next Utterance Prediction}: After hierarchical encoding, we predict the encoding of the next utterance as $b_{t+1}' = \mathrm{FNN}_{1}(e_t)$ where $\mathrm{FNN}_{1}$ is a two-layer feed-forward neural network with layer normalization \cite{layer_norm}. The key objective of DialoGen is to find the historical utterances that are relevant for generating the next utterance. Clearly, there is a requirement for computing a relevance score for all the previous utterances with respect to the next response. But to do so, we need to know the ground-truth response, which is not accessible during prediction time. For this reason, we approximate the next utterance using $b_{t+1}'$. Hence, we need $b_{t+1}'$ to be very close to the utterance-level encoding of the next response i.e. $b_{t+1}$. To ensure that, we introduce a prediction loss ($\mathcal{L}_{pred}$) which is the L1 loss between $b_{t+1}'$ and $b_{t+1}$. 
\begin{equation} \label{eqn:lpred}
     \mathcal{L}_{pred} = \sum_{j=1}^{d} {|{b_{t+1}}_{j} - {b_{t+1}'}_{j}|}
\end{equation} 

\textbf{Relevance Score and Context Vector}: Next, we find the relevance scores for all the previous utterances using $b_{t+1}'$. Let $B = \{b_1, b_2, ..., b_t \} \in \mathbb{R}^{t \times d}$ be all the utterance-level encodings till turn $t$. Then we compute the relevance score as $\alpha^{(t)} = att(B, b_{t+1}') \in \mathbb{R}^{t}$ where $att$ is an additive attention function \cite{nmt}. Finally, we compute a vector representation of the entire context till turn $t$ as $X_t = \sum_{i=1}^{t} \alpha_{i}b_{i}$. To learn a meaningful representation $X_t$, we introduce a Bag-of-Word (BoW) loss defined as,
\begin{equation} \label{eqn:4}
    p_{t} = \mathrm{softmax}(\mathrm{FNN}_{2}(X_t)) \in \mathbb{R}^{|V|}
\end{equation}
\begin{equation} \label{eqn:5}
    \mathcal{L}_{bow} = -\sum_{j=1}^{T} \log p_{{t}_j}
\end{equation}
where $p_{{t}_j}$ is the probability of predicting the $j^{th}$ token in $u_{t+1}$, which is the next utterance in the ground-truth. $T$ is the total number of tokens in $u_{t+1}$, $\mathrm{FNN}_{2}$ is a feed-forward neural network, and $|V|$ is the vocabulary size of BERT tokens. Note that $\mathcal{L}_{bow}$ helps to get a meaningful representation for $X_t$ by actually learning how to combine the previous contexts. Hence, the BoW loss plays an important role in learning the relevance score $\alpha^{(t)}$.

\textbf{Training Objective}: We train the encoder to jointly optimize $\mathcal{L}_{pred}$ and $\mathcal{L}_{bow}$. So, the final loss of the encoder ($\mathcal{L}_{enc}$) is defined as, 
\begin{equation} \label{eqn:6}
    \mathcal{L}_{enc} = \mathcal{L}_{pred} + \mathcal{L}_{bow} 
\end{equation}

\subsection{Decoder}\label{sec:decoder}
The decoder of DialoGen is built on top of pre-trained GPT-2 \cite{gpt2} with a language model head. For a given turn $t$, the decoder first takes $b_{t+1}'$, $\alpha^{(t)}$, and $X_t$ as input from the encoder and selects the top-$k$ historical utterances using $\alpha^{(t)}$. Then a unified representation $Z_t$ is computed combining the past ($X_t$) and predicted ($b_{t+1}'$) contexts. Finally, $Z_t$ is concatenated with the encoding of top-$k$ relevant utterances and fed to GPT-2 to generate the reply.

\textbf{Construction of Decoder Context}: For a given turn $t$, we first combine $X_t$ and $b_{t+1}'$ and compute the unified representation $Z_t$ as,
\begin{equation} \label{eqn:7}
 Z_t = \mathrm{FNN}_3([X_t; b_{t+1}']) \in \mathbb{R}^{d'} 
\end{equation}
where $\mathrm{FNN}_3$ is a feed-forward neural network with layer normalization, and $d'$ is the dimension of GPT-2 embedding. Similar to the encoder, we introduce a bag-of-word loss ($\mathcal{L}_{bow'}$) for the decoder as well. Here, the tokens of $u_{t+1}$ is predicted conditioned on $Z_t$. We use the same method shown in Equations \ref{eqn:4} and \ref{eqn:5} to compute $\mathcal{L}_{bow'}$ except we use $Z_t$ instead of $X_t$. 

In dialogue systems, there is a need to understand the context with a higher level of abstraction to generate the next response~\cite{masknfocus}. In DialoGen, this notion of abstraction is realized through $Z_t$ which is composed of not only historical context ($X_t$) but also the prediction of next utterance ($b_{t+1}'$). The introduction of bag-of-word loss ($\mathcal{L}_{bow'}$) in the decoder helps to learn a meaningful representation for $Z_t$.
 
To include the relevant utterances in the final context, we first choose the top-$k$ utterances based on the relevance scores $\alpha^{(t)}$. Let $R_t$ be the list of top-$k$ relevant utterances in chronological order. We tokenize each utterance in $R_t$ and concatenate them using a special token $[EOS]$ to get token-level encoding $Y_t \in \mathbb{R}^{m \times d'}$ where $m$ is the total number of tokens in $Y_t$. The final context ($C_t$) is defined as the concatenation of $Z_t$ and $Y_t$.
\begin{equation}
    \label{eqn:final}
    C_t = [Z_{t}; Y_{t}] \in \mathbb{R}^{(m+1) \times d'}
\end{equation}
Representing the context in this manner enables to capture not only the entire dialogue history but also helps to focus on the important portions of the relevant utterances via self-attention while generating the response. We consider maximum $N$ tokens from each utterance in $R_t$. Hence, $m$ remains upper bounded by $kN$, where $k$ and $N$ can be set according to the requirement. On the contrary, existing concatenation-based encodings keep on adding previous utterances until the maximum token limit is exceeded. In other words, they use last-$k$ utterances as the context where $k$ may be different for different samples depending on the length of the individual utterances. Selection of the relevant past utterances in $R_t$ and ensuring none of them is left out while forming the context $C_t$ makes the proposed method generalized for long-range context representation.

\textbf{Training Objective}: The GPT-2 model takes $C_t$ as input and generates the next utterance. The language modeling loss ($\mathcal{L}_{LM}$) for generating $u_{t+1}$ given the context $C_t$ is defined as,
\begin{equation} \label{eqn:8}
\mathcal{L}_{LM} = - \sum_{n=1}^{T} \log p(u_{{t+1}_{n}}|u_{{t+1}_{<n}}, C_{t}; \theta)
\end{equation}
where $T$ is the number of tokens in the generated response $u_{t+1}$, and $\theta$ denotes the parameters of the GPT-2 with language model head. We train the decoder to jointly optimize $\mathcal{L}_{bow'}$ and $\mathcal{L}_{LM}$. The final loss of the decoder ($\mathcal{L}_{dec}$) is defined as follows, 
\begin{equation} \label{eqn:9}
    \mathcal{L}_{dec} = \mathcal{L}_{LM} + \lambda*\mathcal{L}_{bow'} 
\end{equation}
where $\lambda$ is a hyper-parameter to set the weightage of the bag-of-word loss.

\section{Experimental Setup}
\subsection{Dataset}
We perform our experiments on DailyDialog \cite{dailydialog} and MultiWOZ 2.1 \cite{multiwoz-2.1} for the generation and understanding tasks, respectively. DailyDialog is a popular open-domain conversational dataset, whereas MultiWOZ is one of the largest datasets for Dialogue State Tracking (DST). Studying the utility of our proposed context representation requires datasets with long-range context dependencies where the next utterance often depends on multiple past utterances in the conversation, which may not be consecutive ones. This is why we use DailyDialog and MultiWOZ, where long-range dependencies can be easily observed. 



\subsection{Implementation Details}
\label{sec:imp}
In the encoder, we use a pre-trained \textit{bert-base-uncased} model having embedding dimension $d=768$ and a stacked GRU with 2 layers i.e. $l=2$. Parameters of the BERT are not updated during training. For the decoder, we finetune pre-trained DialoGPT \cite{dialogpt} instead of vanilla GPT-2. We specifically use \textit{DialoGPT-large} with embedding size $d'=1280$.
\begin{table*}[ht]
\begin{scriptsize}
\centering
\begin{tabular}{c|H|l|rrrr|rr|r|rr|r}
\hline \textbf{ID} & \textbf{Model} & \textbf{Context} & \textbf{Bleu-1} & \textbf{Bleu-2} & \textbf{Bleu-3} & \textbf{Bleu-4} & \textbf{Nist-2} & \textbf{Nist-4} & \textbf{Meteor} & \textbf{Div-1} & \textbf{Div-2} & \textbf{Entropy}\\ \hline
1 & DialoGPT & last-4 & 49.03 & 27.15 & 16.80 & 10.94 & 3.74 &  3.95 & 16.32 & 0.042 & 0.222 & 9.83 \\ 
2 & DialoFlow & all & 48.75 & 26.73 & 16.35 & 10.70 & 3.76 & 3.97 & 16.44 & 0.039 & 0.216 & 9.98 \\
3 & DialogVED & all & 50.50 & 28.95 & 18.38 & 12.29 & 3.94 & 4.18 & 16.90 & 0.037 & 0.204 & 9.82 \\
4 & DialoGen (ours) & top-2 + last-2 &  49.13 & 27.25 & 16.88 & 11.07 & 3.76 & 3.98 & 16.40 & 0.043 & 0.223 & 9.88 \\ \hline
\multicolumn{12}{c}{\textbf{Ablation Study}} \\ \hline
5 & DialoGen (only $Z_t$, no historical utterances) & - & 44.20 & 21.28 & 11.48 & 6.64 & 2.92 & 3.02 & 13.82 & 0.024 & 0.111 & 8.59 \\
6 & DialoGen (only historical utterances, no $Z_t$) & top-2 + last-2& 48.93 & 27.04 & 16.61 & 10.74 & 3.72 & 3.93 & 16.26 & 0.042 & 0.220 & 9.81 \\ \hline
7 & DialoGen & top-4 & 48.39 & 26.51 & 16.24 & 10.54 & 3.65 & 3.85 & 16.04 & 0.041 & 0.214 & 9.84 \\
8 & DialoGen & top-3 + last-1 & 48.80 & 27.02 & 16.70 & 10.97 & 3.71 & 3.93 & 16.22 & 0.042 & 0.218 & 9.86 \\
9 & DialoGen & top-2 + last-2 & 49.13 & 27.25 & 16.88 & 11.07 & 3.76 & 3.98 & 16.40 & 0.043 & 0.223 & 9.88 \\
10 &DialoGen & top-1 + last-3 & 48.98 & 27.09 & 16.70 & 10.95 & 3.72 & 3.94 & 16.28 & 0.041 & 0.216 & 9.86 \\
11 &DialoGen & last-4 & 48.84 & 27.10 & 16.76 & 11.02 & 3.73 & 3.95 & 16.41 & 0.043 & 0.227 & 9.95 \\
\hline
12 & DialoGen (L2) & top-2 + last-2 & 49.08 & 27.15 & 16.69 & 10.87 & 3.74 & 3.95 & 16.25 & 0.041 & 0.216 & 9.85 \\
13 & DialoGen (Cosine) & top-2 + last-2 & 49.03 & 27.05 & 16.51 & 10.57 & 3.72 & 3.92 & 16.32 & 0.040 & 0.211 & 9.78 \\
\hline
\end{tabular}
\caption{\label{tbl:result} Dialogue generation result on DailyDialog dataset
}
\label{table:result}
\end{scriptsize}
\end{table*} 
In the decoder objective, the weight of the bag-of-word loss $\lambda$ is set to 0.5. For decoding, we use beam search with beam width 5, maximum sequence length 40, minimum sequence length 11, and 0.1 length penalty. The same decoding configuration is used to generate the results for the baseline models as well. The rest of the details are provided in Appendix \ref{sec:add_imp}.

Since we are proposing to use top-$k$ relevant utterances, it may happen that the last turn may be excluded. Let's say that we are at turn $t$, and trying to generate the next response for turn $(t+1)$. Then utterance $u_t$ may not be part of the top-$k$ relevant utterances. However, even if $u_t$ may not be important content-wise, it is important to maintain consistency and flow while generating the response. Moreover, note that we are fine-tuning DialoGPT, which is trained using the concatenation of last-$k$ utterances as context. Hence, $u_t$ plays a key role in the generation in DialoGPT. This is why the exclusion of $u_t$ can break the consistency and result in the generation of inconsistent responses. To address this issue, we keep the last-$m$ utterances as part of the context and pick top-$k$ from the remaining previous utterances where $k+m=c_{max}$. In this work, we use $c_{max}=4$, i.e., we restrict ourselves to using up to 4 previous utterances to generate the next response. So, $m=0$ is equivalent to using only top-$k$ relevant utterances, whereas $m=c_{max}$ means only last-$m$ utterances are used as context. For simplicity, we call this model variation as DialoGen with context (top-$k$ + last-$m$). The main result is shown using $k=2$ and $m=2$.

\subsection{Evaluation Metrics}
For the generation task, we use five different metrics - BLEU \cite{bleu}, NIST \cite{nist}, METEOR \cite{meteor}, Diversity \cite{diversity}, and Entropy \cite{entropy}. In general, these metrics struggle to evaluate conversational responses because of the one-to-many nature of dialog \cite{liu-etal-2016-evaluate, metric-survey}. As a result, dialogue generation has to still rely on human evaluation. For the understanding task, we use Joint Goal Accuracy \cite{dstc2, trade} to evaluate DST.

\subsection{Baseline Models}
For DailyDialog, we use the large versions of DialoGPT \cite{dialogpt}, DialoFlow \cite{dialoflow}, and DialogVED \cite{dialogved} as baselines. Note that DialoGen becomes DialoGPT if we remove $Z_t$ and use only last-$k$ utterances as context. This is why we train DialoGPT with last 4 turns as context in order to study the effect of architectural changes in DialoGen. However, no such changes are made in DialoFlow and DialogVED, i.e., both these models use the entire dialogue history as context (truncated by maximum sequence length). The dialogue understanding results on MultiWOZ are shown using SOM-DST \cite{som-dst} and Trippy \cite{trippy}. All the models are trained using the official codes publicly available.

\section{Results}
\subsection{Dialogue Generation (DailyDialog)}
\textbf{Automated Evaluation}: Table \ref{table:result} shows the results of dialogue generation. We have the following observations from the main result (models 1-4). Firstly, DialoGen outperforms DialoGPT in all the metrics. This indicates that the add-on contexts ($Z_t$ and top-2 relevant utterances) have successfully improved the performance of DialoGPT. Secondly, DialoGen performs better than DialoFlow on the BLEU and Diversity but falls short in NIST, METEOR, and Entropy. Thirdly, DialogVED outperforms all the models in BLEU, NIST, and METEOR. However, DialogVED performs worse than all the models in the Diversity and Entropy metric. The root cause of this behavior is the prediction of the next two future tokens while training DialogVED. This results in memorization of n-grams from training data, which negatively impacts the diversity scores during the testing phase. Consequently, there is a boost in the performance of $n$-gram based metrics at the cost of lexical diversity. Later, we show that DialogVED loses to DialoGen on human evaluation, which restates the fact that automated metrics are not reliable for dialogue generation. 

\begin{figure}[t]
     \centering
     \begin{subfigure}[b]{0.235\textwidth}
         \centering
         \includegraphics[width=\textwidth, trim=0 0 200 0,clip]{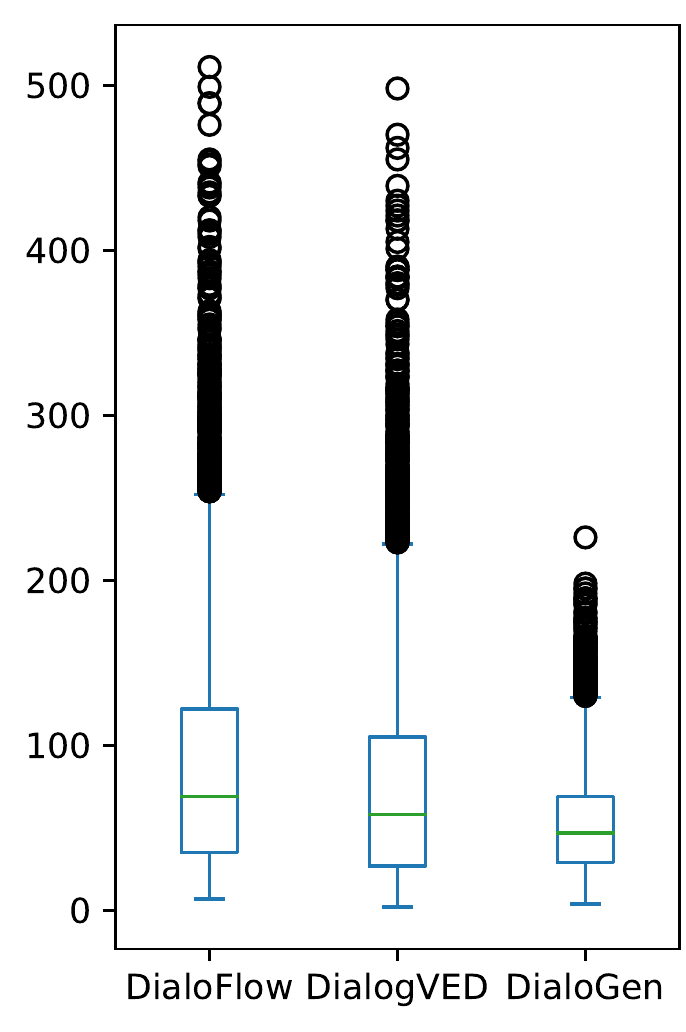}
         \caption{All context}
         \label{fig:box_all}
     \end{subfigure}
     \hfill
     \begin{subfigure}[b]{0.235\textwidth}
         \centering
         \includegraphics[width=\textwidth, trim=200 0 0 0,clip]{tok_len_boxplot.pdf}
         \caption{Context with >4 turns}
         \label{fig:box_g4}
     \end{subfigure}
    \caption{Boxplot of the number of tokens required to represent the dialogue context for different models.}
    \label{fig:box_tok_len}
\end{figure}

Next, we analyze the number of tokens required to represent the dialogue context while computing the results in Table \ref{table:result}. Fig. \ref{fig:box_tok_len} shows the boxplots of the number of tokens consumed by different models. DialoFlow and DialogVED use last-k strategy and consume an average of 89 (max 511) and 77 (max 498) tokens to represent the dialogue history, respectively. In contrast, DialoGen relies on relevant utterances and a high-level abstraction of the entire context ($Z_t$) to meaningfully represent the long-range context compactly. As a result, it consumes an average of 52 tokens (max 226), as shown in Fig. \ref{fig:box_all}. For dialogues with more than four historical utterances/turns in the dataset (around 46\% of the test instances), the average token consumption of DialoFlow, DialogVED, and DialoGen are 143, 125, and 70, respectively (shown in Fig. \ref{fig:box_g4}). 
Therefore, although using a short and compact context representation, DialoGen achieves comparable performance with DialoFlow and DialogVED. This indicates that DialoGen can obtain a representative compression of the dialog context, which, when fed to the decoder, generates responses with similar quality. Furthermore, this compression results in a lesser number of tokens to be processed by the decoder, where each layer spends $O(n^2)$ time for computing self-attention ($n$ being the number of tokens) - thereby reducing the overall memory, compute and power requirements.

\textbf{Ablation Study}: We conduct several experiments to study (a) the impact of $Z_t$ and relevant utterances in the final context representation, (b) the effect of $k$ and $m$  on the automated metrics, and (c) different types of loss functions for next utterance prediction ($\mathcal{L}_{pred}$). In Table \ref{table:result}, model 5 shows the results with context $C_t = Z_t$ whereas model 6 indicates the results with only historical utterances, i.e. $C_t = Y_t$. We can observe that the removal of either of the features has degraded the performance of the original model (shown as model 4). This shows the significance of both $Z_t$ and relevant utterances while constructing the decoder context in Eqn. \ref{eqn:final}. In turn, it also shows the importance of the next utterance prediction and the two bag-of-word losses, which guide the selection of relevant utterances and learning $Z_t$, respectively. Furthermore, the relevant utterance has a stronger influence than $Z_t$ in the final context representation, which aligns with the earlier discussion on achieving state-of-the-art results with concatenation-based encoding. 

Next, models 7-11 in Table \ref{table:result} show the result of DialoGen (top-$k$ + last-$m$) with different values of $k$ and $m$ where $k+m=4$. We observe that including last-$m$ utterances in the relevant set helps improve the model performance. Also, reducing the value of $k$ starts degrading the performance after a certain point. In our case, the best performance is achieved using $k=2$ and $m=2$, which are used to report the main result of DialoGen as well as the human evaluation.

Models 12 and 13 show the result with L2 and Cosine Similarity as the next utterance prediction loss in the DialoGen encoder, respectively. We can observe that L1 (model 4) performs better than both L2 and Cosine Similarity for this purpose. As discussed earlier, there exists a one-to-many mapping from context to dialogue response. This is why L1 is a better loss function for the next utterance prediction since it is robust to outliers.

\textbf{Human Evaluation}: For human evaluation, we randomly picked 300 conversations from DailyDialog test data. For each conversation, we again randomly picked a turn $t$. We displayed the original conversation till turn $t$ and showed generated responses from two models (A and B) to the evaluator, asking for an overall judgment. The evaluators were given four options: i) A is better than B, ii) B is better than A, iii) Both are equally good, and iv) Both are equally bad. We performed this experiment to compare DialoGen with all the other baselines on the same 300 conversations and context points. So, we had a total of 900 response pairs evaluated by 30 humans with moderate inter-annotator agreement (Fleiss' kappa \cite{fleiss} score 0.51). Table \ref{tbl:heval} shows the results of human evaluation. We can observe that DialoGen has a clear edge on human evaluation in comparison to the other baselines.

\begin{table}[t]
\begin{scriptsize}
\centering
\begin{tabular}{|l|r|r|r|r|}
\hline \textbf{Comparison} & \textbf{\%Win} & \textbf{\%Lose} & \textbf{\%Tie} & \textbf{\%Bad} \\ \hline
DialoGen vs DialoGPT & 24.3 & 18.7 & 43.0 & 14.0 \\
DialoGen vs DialoFlow & 30.7 & 22.0 & 35.3 & 12.0 \\
DialoGen vs DialogVED & 31.0 & 24.3 & 34.7 & 10.0 \\
\hline
\end{tabular}
\caption{Human Evaluation on DailyDialog}
\label{tbl:heval}
\end{scriptsize}
\end{table}

\subsection{Dialogue Understanding (MultiWOZ)}
For the dialogue understanding task, we study the effect of utilizing relevant utterances as context in existing Dialogue State Tracking (DST) models. To do so, we first train our DialoGen encoder on the MultiWOZ 2.1~\cite{multiwoz-2.1} dataset for DST. Next, we use the relevance score to compute the context as (top-2 and last-2) and feed it to existing DST models as dialogue history. We experiment with Trippy~\cite{trippy} and SOM-DST~\cite{som-dst}, which are BERT-based DST models having concatenation-based context encoding. We use the official code along with the default settings to train both models and take the average of five runs to report the joint accuracies.
Table \ref{tbl:dst} shows that (top-2+last-2) strategy performs better than last-4 for both models. Moreover, the performance of (top-2+last-2) is close to the performance with all the previous utterances as dialogue history, which correlates with the earlier observations from Table \ref{tbl:result}. Hence, the utilization of relevant utterances can help in dialogue understanding tasks as well.

\begin{table}[b]
\begin{scriptsize}
\centering
\begin{tabular}{|c|l|l|r|}
\hline \textbf{ID} & \textbf{Model} & \textbf{Context Strategy} & \textbf{Joint Acc.} \\ \hline
1 & Trippy & all & 52.97\% \\
2 & Trippy & last-4 & 51.91\% \\
3 & Trippy & top-2 + last-2  & 52.67\% \\ 
\hline
4 & SOM-DST & all & 53.01\% \\
5 & SOM-DST & last-4 & 52.39\% \\
6 & SOM-DST & top-2 + last-2   & 52.82\% \\
\hline
\end{tabular}
\caption{Impact of relevant context on DST}
\label{tbl:dst}
\end{scriptsize}
\end{table}

\begin{table*}[ht]
\centering
\begin{scriptsize}
\begin{tabular}{|T|M|N|N|N|N|N|N|N|N|N|N|N|N|M|}
\hline
\textbf{Turn} & \textbf{Utterance} & $\alpha_1$ & $\alpha_2$ & $\alpha_3$ & $\alpha_4$ & $\alpha_5$ & $\alpha_6$ & $\alpha_7$ & $\alpha_8$ & $\alpha_9$ & $\alpha_{10}$ & $\alpha_{11}$ & $\alpha_{12}$ & \textbf{Model Generated Response}\\ \hline
1 & oh , so many kinds of winter hats . & \underline{1.00} & - & - & - & - & - & - & - & - & - & - & - & yes , there are so many styles of winter hats . \\ \hline
2 & what is your favorite color , miss ? & \underline{0.28} & \underline{0.72} & - & - & - & - & - & - & - & - & - & - & i like red , yellow , blue , black and white . \\ \hline
3 & red . & \underline{0.57} & \underline{0.27} & \underline{0.16} & - & - & - & - & - & - & - & - & - & red is a very nice color . i like it too . \\ \hline
4 & here you are . it 's very attractive . & \underline{0.21} & \underline{0.01} & \underline{0.18} & \underline{0.60} & - & - & - & - & - & - & - & - & thank you . it 's very kind of you . \\ \hline
5 & may I try it on ? & \underline{0.00} & \underline{0.00} & \underline{0.00} & 0.00 & \underline{1.00} & - & - & - & - & - & - & - & go ahead . it fits you well . it 's very fashionable now . \\ \hline
6 & go ahead . & \underline{0.49} & 0.03 & \underline{0.27} & 0.16 & \underline{0.04} & \underline{0.01} & - & - & - & - & - & - & it fits me very well . i like it very much . \\ \hline
7 & is there a mirror around here ? & \underline{0.19} & \underline{0.17} & 0.01 & 0.03 & 0.14 & \underline{0.00} & \underline{0.46} & - & - & - & - & - & i 'm sorry . there is n't one right now . \\ \hline
8 & right over there . & \underline{0.39} & 0.05 & 0.18 & \underline{0.20} & 0.05 & 0.01 & \underline{0.05} & \underline{0.07} & - & - & - & - & oh , i see it now . thank you very much . \\ \hline
9 & does it suit me ? & 0.00 & \underline{0.19} & 0.00 & 0.00 & 0.06 & 0.00 & \underline{0.40} & \underline{0.00} & \underline{0.35} & - & - & - & yes , please sit down . i will show you the way . \\ \hline
10 & yes , you look very nice . & \underline{0.30} & 0.01 & 0.17 & \underline{0.18} & 0.02 & 0.01 & 0.01 & 0.07 & \underline{0.01} & \underline{0.22} & - & - & do you have any idea how much this hat costs ? \\ \hline
11 & how much is it ? & \underline{0.00} & 0.00 & 0.00 & 0.00 & 0.00 & 0.00 & 0.00 & 0.00 & \underline{0.00} & \underline{0.00} & \underline{1.00} & - & it 's \$ 120. but you can get it cheaper . \\ \hline
12 & we 're having a clearance sale . \$ 4 plus tax . & \underline{0.08} & 0.01 & 0.06 & 0.01 & \underline{0.07} & 0.01 & 0.02 & 0.03 & 0.03 & 0.01 & \underline{0.00} & \underline{0.67} & it 's too expensive . i 'll have to pass . \\ \hline
\end{tabular}
\caption{\label{tbl:example} A comprehensive example showing the relevance score and generated response for each turn. The underlined values indicate the turns that were considered to form the decoder context.}
\end{scriptsize}
\end{table*}

\section{Discussions}
\label{sec:discussion}
\subsection{Nature of Context Representation}
In this section, we discuss the nature of context representation in DialoGen. Due to the usage of hierarchical encoding on the encoder side, DialoGen can encode conversations of any length. However, as discussed earlier, hierarchical encoding has its own limitations, and because of that, it cannot outperform concatenation-based methods. To mitigate this issue, we introduce the procedure of finding the previous utterances that are relevant for generating the next response. Our final context (Eqn. \ref{eqn:final}) is represented as the concatenation of the context vector $Z_t$ and encoding of the relevant utterances ($Y_t$). This strategy of context representation has several advantages. Firstly, it is capable of handling arbitrary long conversations. Secondly, it looks only at the exclusive set of relevant turns at the decoding time, which also resolves the long-range dependencies in an interpretable way. Due to this property, it is adaptive to both short and long-range contexts. Hence, the context representation of DialoGen has better generalization than the earlier techniques. Thirdly, since we consider only a limited number of relevant turns, our representation limits the input sequence length (in number of tokens) for the decoder. Consequently, DialoGen can encode conversations of any length compactly while also keeping a check on the overall computational cost of the self-attention operations. With the growing popularity of dialogue systems, we can anticipate datasets with lengthy conversations and an abundance of long-range dependencies. DialoGen is already capable of handling such datasets due to its generalized context representation.

\subsection{Qualitative Analysis and Interpretability}
\label{sec:explain}
As discussed in Eqn. \ref{eqn:final}, the final context of DialoGen is formed by concatenating the relevant utterances with the context vector $Z_t$. The top-$k$ relevant utterances can be interpreted as the extractive summary of the conversation, whereas the context vector $Z_t$ can be viewed as the abstractive summary. In contrast, the concatenation-based strategies use a simple approximation of using only the last-$k$ turns as relevant context, while hierarchical strategies only rely on the abstractive summary. Hence, DialoGen delivers a more meaningful context representation in comparison to earlier methods. In addition to that, the relevance scores make the resolution of long-range dependencies interpretable and give insights into the selection of relevant utterances. We also provide a psycholinguistic analysis of DialoGen architecture in Section \ref{sec:psycho}.

Let us now have a closer look at the actual predictions of DialoGen for a sample conversation. Table \ref{tbl:example} shows the detailed prediction of DialoGen (top-2 + last-2) on a test instance of the DailyDialog dataset. The conversation takes place between a customer (odd turns) and a hat seller (even turns). Firstly, we can observe that the model puts more weight on informative utterances. For example,  while generating the response for Turn 10, maximum weight is assigned to turns 1, 3, 4, and 10, which essentially highlights the key dialogues till Turn 10. Moreover, since Turn 1 is considered as part of the decoder context, the word ``hat'' appeared in the generated response, making it meaningful and consistent. Also, the predicted response expresses the same thing as the actual response given in Turn 11. In Turns 5 and 11, we can observe that a question has been asked, i.e., the next speaker is expected to give an answer to that. The model is able to capture it by putting maximum weight (almost 1) on the questions asked at Turns 5 and 11, respectively.

One point to remember is that the relevance scores are not trained explicitly using annotated data. Rather, they are learned as part of the training procedure while maximizing the prediction of bag-of-words of the next utterance (Eqn. \ref{eqn:5}). Because of that, it may not always output the absolute best scores, but the score could be considered as a soft indicator of the importance of a given utterance towards generating the next response.

\subsection{Psycholinguistic Analysis of DialoGen}
\label{sec:psycho}
The mechanism through which a human generates responses in a multi-turn conversation can be divided into three major steps, each of which is a cognitive process: a) perception of the other speaker’s turn, b) planning of one’s own turn, and c) production of the speech or response \cite{speech-plan}. In this sequence, the first process deals with perception, where we carefully listen to the other speaker and try to understand the message with reference to past utterances and other relevant contexts. The human brain is able to analyze and connect all the pieces of information very fast using ``quick-and-dirty'' shortcuts \cite{berkum}. The second process i.e. speech planning, is a multifaceted process central to response generation \cite{speech-plan}. Studies have indicated that humans often start planning their responses while listening to other people \cite{speech-plan2}. In the final process i.e. Generation, our thoughts are translated into words/sentences to generate a response based on the speech plan.

Let us now analyze the extent to which DialoGen matches the above-mentioned procedure of human conversation. Conceptually, DialoGen contains all three cognitive processes involved in multi-turn human conversation. Firstly, the resolution of long-range dependencies and focusing on the relevant previous utterances in DialoGen resembles the first process of understanding and connecting the relevant information. Secondly, DialoGen computes a high-level abstraction of the conversation context ($Z_t$) using the understanding of the historical context ($X_t$) and prediction of the next utterance ($b_{t+1}'$). Note that $Z_t$ is learned using bag-of-word loss to predict the words of the next utterance. The final context is constructed as the concatenation of $Z_t$ and the top-$k$ relevant utterances. This procedure of context encoding can be perceived as the second process of speech planning. Finally, DialoGen uses a language model to generate the response conditioned on the final context, which is similar to the third process of speech production. Hence, there exists a certain degree of similarity between the architecture of DialoGen and the cognitive procedure of human conversation.

\section{Conclusion}

In this work, we analyze the usefulness and limitations of both concatenation-based and hierarchical context encodings for dialog context representation. To take advantage of both methods, we propose a novel conversation generation framework, DialoGen, with a generalized context representation that is adaptive to long-range dependencies. We represent dialogue context as a combination of a context vector and relevant utterances. DialoGen achieves comparable performance to state-of-the-art models on the DailyDialog dataset, regardless of its short and compact representation of dialogue history. Furthermore, it shows better performance than the baseline models in human evaluation. We also observe similar behavior in DST when relevant dialogue history is applied to existing DST models. Finally, we discuss the generalizability and interpretability of the proposed method with a comprehensive example. We also provide a psycholinguistic analysis of DialoGen. In future work, we want to explore the application of our proposed context representation on knowledge or persona-grounded conversation and other dialogue-related complex tasks and datasets.

\bibliography{mybib}
\bibliographystyle{acl_natbib}



\appendix

\section{Appendix}
\label{sec:appendix}

\begin{table}[t]
\centering
\begin{scriptsize}

\begin{tabular}{|p{1cm}|c|r|r|r|r|r|}
\hline \textbf{Dataset} & \textbf{Type} & \textbf{\#Dialogues} & \textbf{\#Turns} & \textbf{T\textsubscript{max}} & \textbf{T\textsubscript{min}} & \textbf{T\textsubscript{avg}} \\ \hline
\multirow{3}{*}{DailyDialog} &
Train & 11118 & 87170 & 35 & 2 & 7.84\\
&
Dev & 1000 & 8069 & 31 & 2 & 8.07\\
&
Test & 1000 & 7740 & 26 & 2 & 7.74\\
\hline
\multirow{3}{*}{MultiWOZ} &
Train & 8420 & 56668 & 22 & 1 & 6.73 \\
&
Dev & 1000 & 7374 & 17 & 2 & 7.37 \\
&
Test & 999 & 7368 & 18 & 2 & 7.37 \\ \hline

\end{tabular}
\caption{Basic statistics of DailyDialog and MuliWOZ dataset. T\textsubscript{max}, T\textsubscript{min}, and T\textsubscript{avg} indicates maximum, minimum, and average dialogue turns.}
\label{tbl:stat}
\end{scriptsize}
\end{table}

\subsection{Dataset details and pre-processing}
The basic statistics of DailyDialog \cite{dailydialog} and MultiWOZ 2.1 \cite{multiwoz-2.1} datasets are shown in Table \ref{tbl:stat}. Data pre-processing for both DailyDialog and MultiWOZ datasets is minimal. All the dialogues are transformed into lowercase texts. While tokenizing the utterances for BERT/GPT-2, we consider a maximum of 64 tokens from each tokenized text i.e. tokenized texts having more than 64 tokens are truncated.

\subsection{Multi-reference data for DailyDialog}
To improve the quality of automatic evaluation for DailyDialog, \citet{multi-ref} augmented the test set of DailyDialog with multiple references. To be more specific, four reference responses are augmented in addition to the original response in the test data. So, all five responses are used as references during the automatic evaluation.

\begin{table}[b]
\centering
\begin{small}
\begin{tabular}{|l|l|}
\hline
\textbf{Turn} & \textbf{Utterance} \\ \hline
1 & oh , so many kinds of winter hats . \\ \hline
2 & what is your favorite color , miss ?\\ \hline
3 & red . \\ \hline
4 & here you are . it 's very attractive . \\ \hline
5 & may i try it on ? \\ \hline
6 & go ahead . \\ \hline
7 & is there a mirror around here ? \\ \hline
8 & right over there . \\ \hline
9 & does it suit me ? \\ \hline
10 & yes , you look very nice . \\ \hline
11 & how much is it ? \\ \hline
\end{tabular}
\caption{\label{tbl:example3} DailyDialog multi-reference data snippet. }
\end{small}
\end{table}

\subsection{Post-processing of generated dialogues}
The reference data of DailyDialog follows a specific format. Table \ref{tbl:example3} shows the formatting of the same example shown in Table \ref{tbl:example2}. We can observe that apart from being lowercase, there are a few subtle formatting styles that the texts follow. For example, ``,'' is both preceded and succeeded by a space. Similarly, ``.'' is preceded by a space. Moreover, words like ``it 's'', ``do 'nt'' have particular formatting. Note that a space is added before ', which is not common. However, most of these issues are resolved by tokenizing the text using a word tokenizer followed by concatenating the tokenized text with spaces. In this work, we use the NLTK library for this purpose. However, this simple word tokenization trick is not sufficient to match the reference format completely. For example, the formatting of words like ``it 's'' and ``do 'nt'' cannot be fully achieved using this trick. To address the issue, we manually found some frequently occurring mismatch patterns and applied a regex-based transformation to further approximate the reference format. These conversions are applied for all the generated conversations, including the baselines. All the automated evaluation results are computed using the converted texts. Due to this reason, results of the baseline models on DailyDialog have a minor deviation from the results reported in the original papers. The post-processing script is shared with the code.

\subsection{Additional Implementation Details}
\label{sec:add_imp}
We implemented DialoGen using PyTorch and Huggingface \cite{huggingface} libraries in Python 3.8. All the experiments are performed on a single device of Nvidia DGX server with 32GB of memory. The number of parameters in the encoder and the decoder is 33M and 840M, respectively. We use dropout ratio of 0.2 and AdamW \citep{adamw} optimizer with adam's epsilon 1e-8. For the encoder, we use a learning rate of 5e-4 and maximum training epochs of 30 while the same values are set to 1e-5 and 10 respectively for the decoder. The encoder and decoder are trained separately as the end-end modeling is not straightforward due to the selection of relevant utterances. The best model is selected based on minimum validation loss. Our current implementation of DialoGen is not parallelizable, so the batch size is fixed to 1. However, we use gradient accumulation and update the parameters after the training of every four conversations, which makes the effective batch size around 28. The average training time of the encoder and decoder is 2.5 hours and 24 hours, respectively. As mentioned in Section \ref{sec:imp}, we do not update the parameters of the BERT model. The final layer of a pre-trained BERT model is biased toward the pre-training tasks. This is why we consider the second-to-last layer of the BERT embeddings instead of [CLS], which is a reasonable sweet spot for using BERT embeddings without fine-tuning \cite{bert}.

The loss values of the encoder and decoder of DialoGen are shown in Tables \ref{tbl:enc_loss} and Table \ref{tbl:dec_loss}, respectively. The automatic metrics for dialogue generation are computed following the evaluation of DSTC7 Task 2~\footnote[1]{\href{https://github.com/mgalley/DSTC7-End-to-End-Conversation-Modeling/tree/master/evaluation/src}{github.com/mgalley/DSTC7-End-to-End-Conversation-Modeling/tree/master/evaluation/src}}. The end-end modeling of DialoGen is not straightforward. This is because of the relevant context selection step in the decoder. In this work, we avoided this problem by training the encoder and decoder separately. However, end-end modeling will not be a problem if the annotations of relevant utterances to generate the next response are already available in the dataset. 

\begin{table}[t]
\centering
\begin{scriptsize}

\begin{tabular}{|c|l|r|r|r|}
\hline \textbf{Data} & \textbf{Type} & \textbf{Total Loss} & \textbf{BoW Loss} & \textbf{L1 Loss} \\ \hline
\multirow{3}{*}{DailyDialog} &
Train & 4.81 & 4.61 & 0.19\\
&
Validation & 5.55 & 5.35 & 0.20 \\
&
Test & 5.57 & 5.37 & 0.20 \\
\hline
\multirow{3}{*}{MultiWOZ} &
Train & 4.74 & 4.58 & 0.16\\
&
Validation & 5.01 & 4.84 & 0.17 \\
&
Test & 4.98 & 4.81 & 0.17 \\
\hline

\end{tabular}
\caption{Encoder loss on DailyDialog and MultiWOZ dataset}
\label{tbl:enc_loss}
\end{scriptsize}
\end{table}

\begin{table}[t]
\centering
\begin{scriptsize}

\begin{tabular}{|l|r|r|r|}
\hline \textbf{Type} & \textbf{Total Loss} & \textbf{LM Loss} & \textbf{BoW Loss} \\ \hline
Train & 4.11 & 1.31 & 5.61\\
Validation & 4.98 & 2.08 & 5.79 \\
Test & 5.00 & 2.08 & 5.83 \\
\hline
\end{tabular}
\caption{Decoder loss on DailyDialog dataset}
\label{tbl:dec_loss}
\end{scriptsize}
\end{table}

\begin{table*}[t]
\begin{scriptsize}
\begin{tabular}{|T|c|K|N|N|N|N|N|N|N|N|N|N|N|N|N|N|N|}
\hline
\textbf{Turn} & \textbf{Speaker} & \textbf{Utterance} & $\alpha_1$ & $\alpha_2$ & $\alpha_3$ & $\alpha_4$ & $\alpha_5$ & $\alpha_6$ & $\alpha_7$ & $\alpha_8$ & $\alpha_9$ & $\alpha_{10}$ & $\alpha_{11}$ & $\alpha_{12}$ & $\alpha_{13}$ & $\alpha_{14}$ & $\alpha_{15}$ \\ \hline
1 & User & I need a train to stansted airport that leaves on Sunday. & \underline{1.00} & - & - & - & - & - & - & - & - & - & - & - & - & - & - \\ \hline
2 & System & Did you have a time you would like to arrive or leave? & - & - & - & - & - & - & - & - & - & - & - & - & - & - & - \\ \hline
3 & User & I need to arrive by 14:30. & \underline{0.65} & \underline{0.21} & \underline{0.14} & - & - & - & - & - & - & - & - & - & - & - & - \\ \hline
4 & System & TR1668 will arrive at 14:08, would that work for you? & - & - & - & - & - & - & - & - & - & - & - & - & - & - & - \\ \hline
5 & User & That is perfect. I would like to make a booking for 6 people please. & \underline{0.02} & 0.00 & \underline{0.05} & \underline{0.53} & \underline{0.40} & - & - & - & - & - & - & - & - & - & - \\ \hline
6 & System & Booking was successful, the total fee is 48.48 GBP payable at the station. Your feference number is HF03UG02. Do you need assistance with anything else? & - & - & - & - & - & - & - & - & - & - & - & - & - & - & - \\ \hline
7 & User & I need to eat too & \underline{0.00} & 0.00 & 0.00 & 0.00 & \underline{0.00} & \underline{0.00} & \underline{1.00} & - & - & - & - & - & - & - & - \\ \hline
8 & System & What type of restaurant and price range are you looking for? & - & - & - & - & - & - & - & - & - & - & - & - & - & - & - \\ \hline
9 & User & I'd like Catalan food. It needs to be in the centre and be expensive.  & 0.00 & 0.00 & 0.00 & 0.00 & \underline{0.00} & 0.00 & \underline{0.03} & \underline{0.00} & \underline{0.97} & - & - & - & - & - & - \\ \hline
10 & System & I'm sorry, there aren't any restaurants like that. Would you like something else? & - & - & - & - & - & - & - & - & - & - & - & - & - & - & - \\ \hline
11 & User & What about one that serves european food in the same side? & 0.00 & 0.00 & 0.00 & 0.00 & 0.00 & 0.00 & \underline{0.01} & 0.00 & \underline{0.09} & \underline{0.02} & \underline{0.88} & - & - & - & - \\ \hline
12 & System & There are three european restaurants in the center of town. Would you like me to pick one? & - & - & - & - & - & - & - & - & - & - & - & - & - & - & - \\ \hline
13 & User & Yes please do and then make me a reservation for 6 people at 10:15 on a Sunday. & 0.00 & 0.00 & 0.02 & 0.02 & \underline{0.08} & 0.01 & 0.01 & 0.00 & 0.04 & 0.00 & \underline{0.11} & \underline{0.08} & \underline{0.61} & - & - \\ \hline
14 & System & You have a table booked for Eraina and the reference number is T7ZSP58S. Is there anything else I can do for you? is there anything else i can do for you ? & - & - & - & - & - & - & - & - & - & - & - & - & - & - & - \\ \hline
15 & User & No, that's all. Thanks! Goodbye. & \underline{0.00} & 0.00 & 0.00 & 0.00 & 0.00 & 0.00 & \underline{0.01} & 0.00 & 0.00 & 0.00 & 0.00 & 0.00 & 0.00 & \underline{0.00} & \underline{0.99} \\ \hline
16 & System & Have a great day & - & - & - & - & - & - & - & - & - & - & - & - & - & - & - \\ \hline
\end{tabular}
\caption{\label{tbl:ex_multiwoz} \small A comprehensive example from the MultiWOZ dataset showing the relevance score for each user turn. The underlined values indicate the turns that were considered to form the dialogue history using (top-2 + last-2) strategy.}
\end{scriptsize}
\end{table*}

\subsection{Analysis of Relevance Score on MultiWOZ datset}
In this section, we analyze the relevance score of the DiaoloGX encoder trained with the MultiWOZ dataset. Table \ref{tbl:ex_multiwoz} shows the predicted relevance score of the encoder on a test instance from the MultiWOZ dataset. Note that relevance scores predicted by this DialoGX encoder have been used to form the (top-2 + last-2) context of model 3 (Trippy) and model 6 (SOM-DST) shown in Table \ref{tbl:dst}. The sample conversation shown in Table \ref{tbl:ex_multiwoz} is basically task-oriented, where a user converses with the system agent to book a train followed by a restaurant. In MultiWOZ, dialogue state predictions are made after each user turn. This is why we show the relevance scores only for the user turns. The score signifies the importance of the previous turns to generate the next system response.

Let us now analyze the relevance scores shown in Table \ref{tbl:ex_multiwoz}. Firstly, we can observe that the relevance score of the current turn is significant for all the user turns. This shows that the model is capable of detecting the importance of the last user turn, which aligns with the conversations of the MultiWOZ dataset. Secondly, the model is able to understand the context switches. In Table \ref{tbl:ex_multiwoz}, there are two context switches (Turns 7 and 15). In both cases, the model is able to put approx. 1 weightage on the current turn and nearly 0 on the rest of the turns. Thirdly, in turn 11, we can observe that the user is referring to the word ``centre'' mentioned in turn 9. The model is able to capture this co-reference by putting the second highest (0.09) score on Turn 9. However, we reiterate that the relevance scores are not trained explicitly using annotated relevance scores. Due to this reason, it is better to consider the relevance score as a soft indicator of the importance of a given utterance in generating the next response.

\begin{figure*}[ht]
    \begin{center}
        \includegraphics[scale=0.7, trim=20 250 20 17, clip]{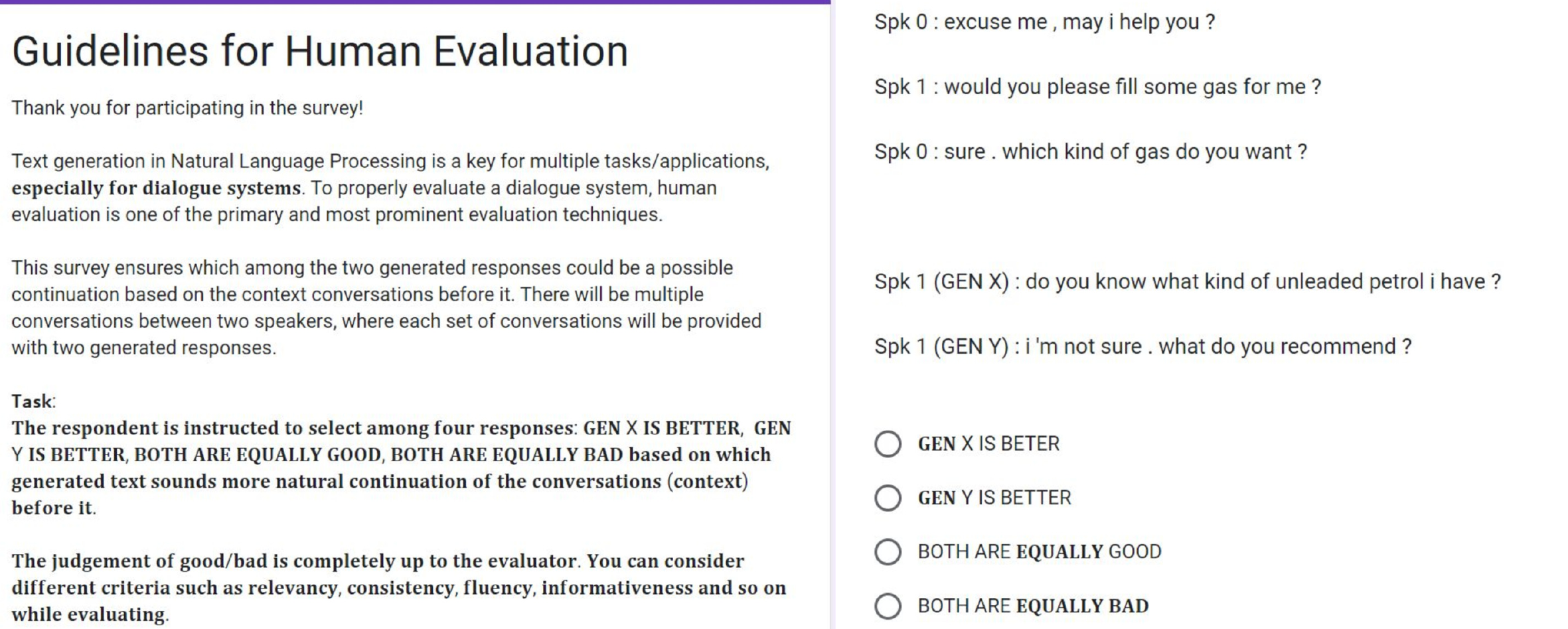}
    \caption{Huaman evaluation form}
    \label{fig:he}
    \end{center}
\end{figure*}

\subsection{Additional Details on Human Evaluation}
We conducted the human evaluation using 30 engineering students proficient in English. The instructions and a sample of the human evaluation form is shown in Fig. \ref{fig:he}. The source data of the human evaluation forms are shared with the code repository. We asked the evaluators to give an overall judgment instead of evaluating for different attributes like relevancy, consistency, informativeness, fluency, etc. Evaluating each dialogue attribute is subjective and requires strong inter-annotator agreement. Asking for a single opinion made the human evaluation process simple, fast, and less confusing, although restricting the analysis of fine-grained aspects of generated dialogues. All the generated dialogues were pretty fluent (due to our usage of strong baselines), which was also a general comment from the human evaluators. Thus, fluency can be ignored as the deciding attribute in Table \ref{tbl:heval}. 

\end{document}